\documentclass[runningheads]{llncs}
\usepackage[T1]{fontenc}
\usepackage{graphicx}
\usepackage{amsmath,amsfonts}
\usepackage{algorithmic}
\usepackage{dsfont}
\usepackage{bbm}

\usepackage{multirow}

\usepackage[ruled,vlined]{algorithm2e}
\usepackage{float}
\usepackage{booktabs}
\usepackage{wrapfig}
\usepackage[misc]{ifsym} % Letter for corresponding author
\usepackage{enumitem}
\begin{document}
%

% \title{Contribution Title\thanks{Supported by organization x.}}
\title{SCME: A Self-Contrastive Method for Data-free and Query-Limited Model Extraction Attack\thanks{This paper was accepted by ICONIP 2023.}}

\titlerunning{SCME}
\author{Renyang Liu\inst{1}\orcidID{0000-0002-7121-1257} \and
Jinhong Zhang\inst{1}\orcidID{0000-0002-9906-3508} \and
Kwok-Yan Lam\inst{2}\orcidID{0000-0001-7479-7970} \and
Jun Zhao\inst{2}\orcidID{0000-0002-3004-7091} \and
Wei Zhou\inst{\textsuperscript{1,~\Letter}}  \orcidID{0000-0002-5881-9436}}
% %
\authorrunning{Renyang Liu et al.}

\institute{Yunnan University, Kunming, China\\
\email{\{ryliu,jhnova\}@mail.ynu.edu.cn,zwei@ynu.edu.cn} \\ 
\and
Nanyang Technological University, Singapore \\
\email{\{kwokyan.lam,junzhao\}@ntu.edu.sg}}

% Submission 837

%
\maketitle              % typeset the header of the contribution
\begin{abstract}
Previous studies have revealed that artificial intelligence (AI) systems are vulnerable to adversarial attacks. Among them, model extraction attacks fool the target model by generating adversarial examples on a substitute model. The core of such an attack is training a substitute model as similar to the target model as possible, where the simulation process can be categorized in a data-dependent and data-free manner. Compared with the data-dependent method, the data-free one has been proven to be more practical in the real world since it trains the substitute model with synthesized data. However, the distribution of these fake data lacks diversity and cannot detect the decision boundary of the target model well, resulting in the dissatisfactory simulation effect. Besides, these data-free techniques need a vast number of queries to train the substitute model, increasing the time and computing consumption and the risk of exposure. To solve the aforementioned problems, in this paper, we propose a novel data-free model extraction method named SCME (Self-Contrastive Model Extraction), which considers both the inter- and intra-class diversity in synthesizing fake data. In addition, SCME introduces the Mixup operation to augment the fake data, which can explore the target model's decision boundary effectively and improve the simulating capacity. Extensive experiments show that the proposed method can yield diversified fake data. Moreover, our method has shown superiority in many different attack settings under the query-limited scenario, especially for untargeted attacks, the SCME outperforms SOTA methods by 11.43\% on average for five baseline datasets.

\keywords{Adversarial Attacks, Model Extraction Attacks, Black-Box Attacks, Model Robustness, Information security.}

\end{abstract}

\section{Introduction}
% \vspace{-10pt}
Recently, Trusted AI, which contains fairness, trustworthiness and explainability, 
\begin{wrapfigure}{r}{6cm}
% \vspace{-20pt}
    % \setlength{\abovecaptionskip}{-0.1cm}
    \setlength{\belowcaptionskip}{-0.8cm} 
    \centering
    \includegraphics[width=0.5\textwidth]{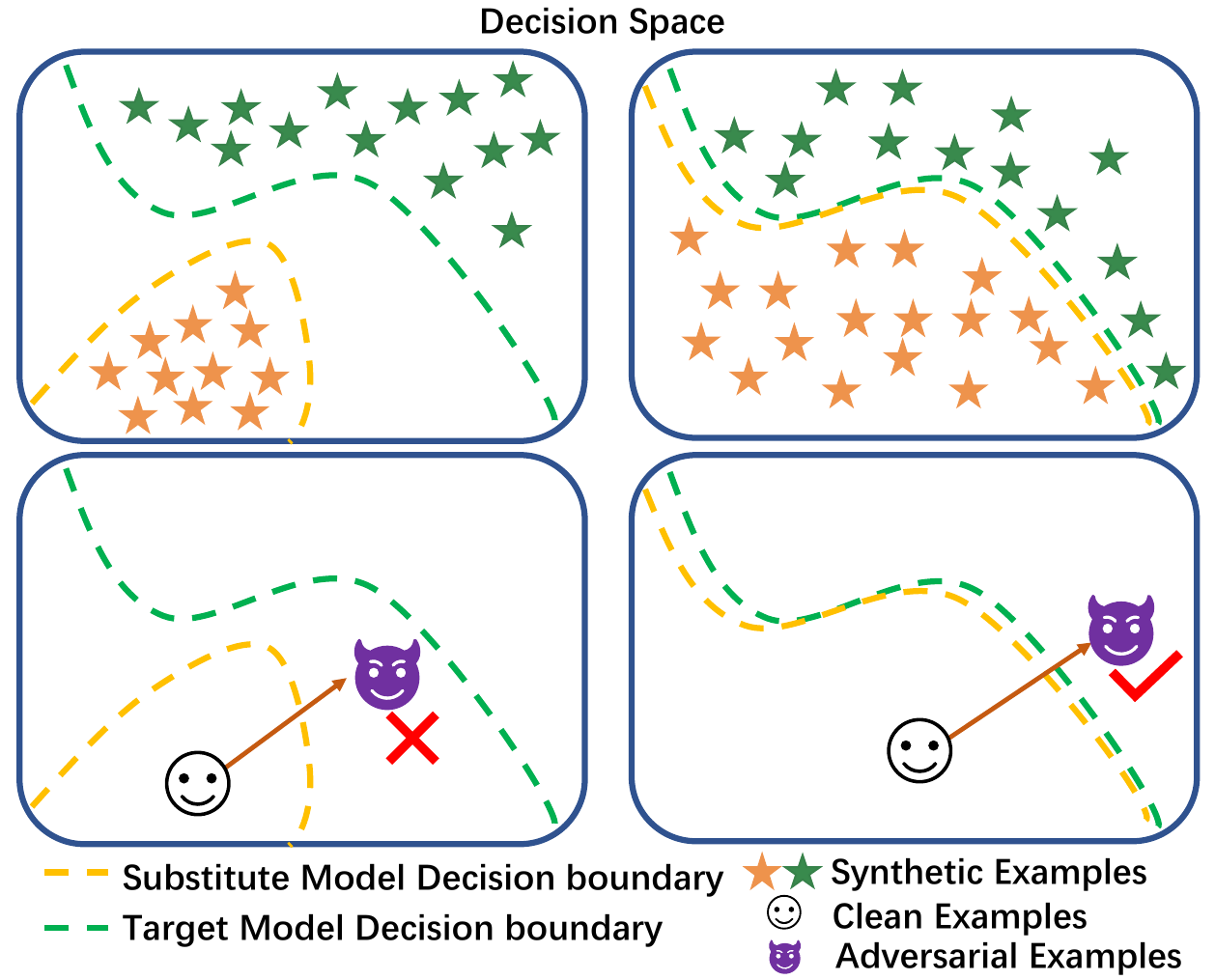}
    \caption{Synthetic example distribution, decision boundaries and whether the attack is successful. \textbf{Top left:} bad synthetic example distribution failed to fit the target model decision boundary. \textbf{Bottom left:} unfitting of the decision boundary leads to attack failure. \textbf{Top right:} good example distribution and decision boundary fit. \textbf{Bottom right:} good decision boundary.}
    \label{fig:Decision}
\end{wrapfigure}
has received increasing attention and plays an essential role in the AI development process. The security of the AI models, however, is being doubted and has bought concerns in academia and industry. A lot of research has shown that AI models (including Machine Learning (ML) models and Deep Learning (DL) models) are vulnerable to adversarial examples \cite{DBLP:conf/asru/DemuynckT13}, which are crafted by adding a virtually imperceptible perturbation to the benign input but can lead the well-trained AI model to make wrong decisions. For example, in the physical world, the attackers can maliciously alter traffic signs by sticking a small patch \cite{DBLP:conf/aaai/LiuLFMZXT19}, changing the content style \cite{DBLP:conf/cvpr/DuanM00QY20} and shooting a laser on it \cite{DBLP:conf/cvpr/DuanMQCYHY21}. Although these modifications do not affect human senses, but can easily trick autonomous vehicles. Therefore, it is imperative to devise effective attack techniques to identify the deficiencies of AI models beforehand in security-sensitive applications \cite{DBLP:journals/tdsc/LiuP21}.

Existing adversarial attack methods on DL models can be categorized into white-box attacks and black-box attacks. In the white-box settings, the attackers can access the whole information of the target model, including weights, inner structures and gradients. In contrast, in the black-box one, the attackers have no permission to access the models' details but the final output \cite{DBLP:conf/iclr/KurakinGB17a,TI-FGSM}. With such rules, it is clear that black-box attacks are more challenging but practical in the physical world, where the attacker lacks details of the target models. To attack DL models in black-box settings more effectively, model extraction attacks have been proposed \cite{DBLP:conf/uss/TramerZJRR16}, which is implemented by training a substitute model and generating adversarial examples on such model to attack the target model successfully.

Most of the previous model extraction attacks \cite{DBLP:conf/ccs/PapernotMGJCS17,Knockoff} concentrates on training the substitute model by querying the target model with real data, called data-dependent model extraction. However, it is infeasible to the physical world, where the adversary can not access the models' training data. As the counterpart, the data-free model extraction attack solved this problem by synthesizing fake data \cite{DAST,FE-DAST}. In this scenario, the attackers use generators to synthesize the fake data to train substitute models. For launching attacks with a high success rate, as shown in Fig. \ref{fig:Decision}, the decision boundary of the substitute model should maintain a very high similarity to the target black-box model. Besides, training the substitute model with synthetic data is challenging to the problem of \textit{How to generate valuable synthetic data for the substitute model training?} Generally speaking, the synthetic examples should have the following two properties: 1) \textbf{inter-class diversity} and 2) \textbf{intra-class diversity}. The inter-class diversity means that the synthetic examples' categories classified by the target model should contain all the expected classes, while the intra-class diversity indicates that the examples should differ from each other, even they belong to the same category. However, existing methods \cite{DAST,FE-DAST,EBFA,Del} still suffer from the following two challenges: 1) They only consider inter-class diversity but ignore intra-class diversity, resulting in synthetic data not serving as well as real data. 2) The other is that the query data are generated in the substitute model's training process. However, once the substitute model is not well-trained, it can hardly provide the effective target model's decision information.

To solve these challenges mentioned above, in this paper, we propose a novel data-free model extraction method, named \textbf{S}elf-\textbf{C}ontrastive \textbf{M}odel \textbf{E}xtraction (SCME for short). SCME introduces the idea of contrastive learning \cite{DBLP:conf/icml/ChenK0H20} and proposes a self-contrastive mechanism to guide the training of the generator. Specifically, we design a self-contrastive loss to enlarge the distance of the substitute model's latent representation. Benefiting from this, the generator will be encouraged to synthesize more diversified fake data. Furthermore, SCME introduces the Mixup operation to interpolate two random images into a single one to build the query examples, which can improve the efficiency of the substitute model in learning the target model's decision boundaries in a model-independent manner. Extensive experiments illustrate SCME can synthesize fake data with diversity and improve the attack performance. Our contributions are summarized as follows:
\begin{itemize}[leftmargin=*, itemsep=1pt, topsep=1pt, parsep=0pt]
    \item We propose a novel data-free model extraction method, called SCME, to generate efficient fake data for the substitute model training under query-limited settings.   
    \item We use a self-contrastive mechanism to guide the generator to synthesize the fake data with inter- and intra-class diversity to help the substitute model imitate the target model efficiently.
    \item We introduce the Mixup into SCME, which can build fake data in a model-independent manner, to detect decision boundaries of the target model effectiveness and further help the imitating processing.
    \item Extensive empirical results show the SCME's superiority in the synthetic diversified fake data and the adversarial examples' attack performance in query-limited situations. 
\end{itemize}

\section{Related Work}
\label{Sec:related}
Previous researches contend that the DL models are sensitive to adversarial attacks, which can be classified into white-box and black-box. In white-box settings, the attackers can generate adversarial examples with a nearly 100\% attack success rate because they can access the target model. The black-box attack, however, is more threatening to the DL models in various realistic applications because they do not need the models' details. Among them, the model extraction attack \cite{DBLP:journals/tetci/YuanDZLW22} has received much attention recently due to its high attack performance.

The success of model extraction attacks relies heavily on adversarial examples' transferability, which means the adversarial examples generate on model $A$ can also attack model $B$ successfully. To implement such an attack, the attacker first trains a surrogate local model by simulating the target models' output. When the surrogate model is well-trained, it will have the same decision boundary as the target model, i.e., output the same results for the same input; this imitation process is called model extraction. However, due to the data bias between the query data used for surrogate model training and the real data used for the target model training, creating a valid query dataset is the crucial point of model extraction attacks. Papernot et al. \cite{DBLP:conf/ccs/PapernotMGJCS17} first used adversarial examples to query the target model for model extraction. However, due to the surrogate model is not well-trained, the adversarial examples generated on it cannot perform well in the imitating process. Orekondy et al. \cite{Knockoff} propose the Knockoff to try to find valid query examples in a huge dataset, e.g., ImageNet \cite{DBLP:conf/nips/KrizhevskySH12}, and adopt an adaptive strategy in the extraction process. Zhou et al. \cite{DAST} proposed DAST, which is the first work to use a generator-based data-free distillation technique in knowledge distillation for model extraction. Later, subsequent studies have improved this approach to achieve better results \cite{Del}. However, the generator-based approach cannot obtain sufficient supervised information as in white-box knowledge distillation, leading to a huge number of queries and low attack results.

Therefore, the block-box attack with adversarial examples' transferability poses the request to guarantee that the local model is highly similar to the target model. To achieve this goal, we know from previous studies that model extraction can steal the target model from a decision boundary perspective, even in a data-independent way. However, the previous data-free works can not guarantee the synthetic data's diversity and need a massive number of queries to the target model. Hence, we are well-motivated to develop a better model extraction strategy adapted to data-free settings for carrying out attacks with high performance. Besides, it can improve the diversity of the generated fake data to be suitable for query-limited settings.  

\section{PRELIMINARY}
\label{sec:preliminary}

\subsection{Adversarial Attack}
Given a classifier $\mathcal{F}(\cdot)$ and an input $ x $ with its corresponding label $ y $, we have $ \mathcal{F}(x)=y $. The adversarial attack aims to find a small perturbation $\delta$ added to $x$, so the generated input $x'$ misleads the classifier's output. The perturbation $\delta$ is usually constrained by $L_p$-norm $(p=1,2, ..., \infty)$, i.e., $\Vert \delta \Vert_{p} \le \epsilon$. Then, the definition of adversarial examples $x'$ can be written as:
\begin{equation}
    \begin{aligned}
        &\mathcal{F}(x') \neq {y}_{true}, 
        &s.t.~\Vert x'-x \Vert_{p} \le \epsilon,\\
    \end{aligned}
\end{equation}
where $\epsilon$ is the noise budget, $y_{true}$ is the ground-truth label of example $x$.

\begin{figure}[htp]
    \vspace{-10pt}
    \setlength{\abovecaptionskip}{-0.0cm}
    \setlength{\belowcaptionskip}{-0.1cm} 
    \includegraphics[width=\textwidth]{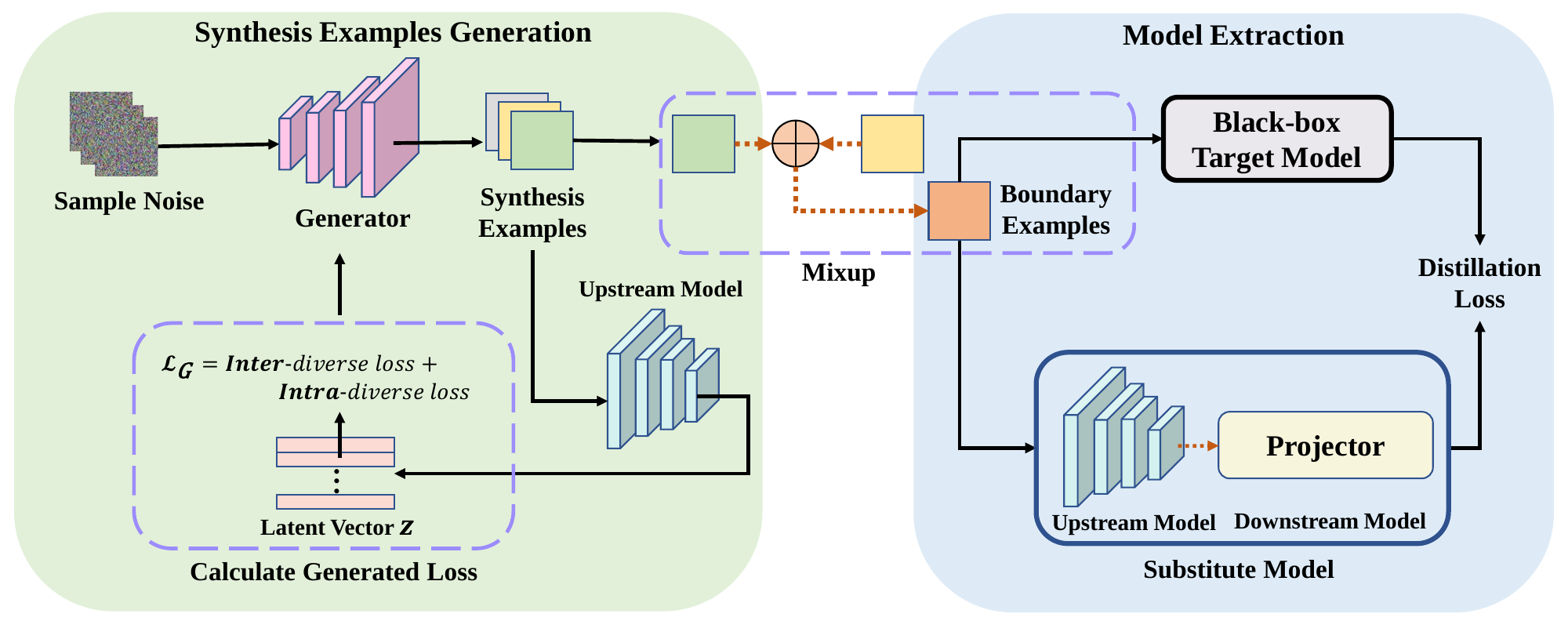}
    \caption{Framework of SCME, where $\mathcal{L}_G $ consisting of inter- and intra-diverse loss.}
    \label{fig:framework}
    \vspace{-10pt}
\end{figure}

\subsection{Contrastive Learning}
Contrastive learning models usually consist of two portions: self-supervised training in the upstream network and supervised fine-tuning in the downstream network. The upstream network $f(\cdot)$ aims to maximize the paired instance by augmenting the same data in different ways in the learned latent space while minimizing the agreement between different instances. Given a batch of examples $\{x_{N}\}$ without ground-truth labels, the random data transformation $\mathcal{T}$ takes each example $x$ in $\{x_{N}\}$ to a paired augmented data copies $x_i$ and $x_j$, resulting in $2N$ augmented examples. The trained upstream network $f(\cdot)$ encodes the paired copies to latent vectors $z_i$ and $z_j$. n SimCLR \cite{DBLP:conf/icml/ChenK0H20}, the contrastive loss can be formulated as:
\begin{equation}
    \ell_{i, j}=-\log \frac{\exp \left(\operatorname{sim}\left(\boldsymbol{z}_{i}, \boldsymbol{z}_{j}\right) / \tau\right)}{\sum_{k=1}^{2 N} \mathbbm{1}_{[k \neq i]} \exp \left(\operatorname{sim}\left(\boldsymbol{z}_{i}, \boldsymbol{z}_{k}\right) / \tau\right)},
\end{equation}
where the $z_i$ and $z_j$ are the latent vectors of positive augmented examples, and $z_k$ indicts the latent vector of negative examples from a different class. The $sim(\cdot)$ is a similarity function, such as cosine similarity loss, $\mathbbm{1}$ is the indicator function, and $\tau$ is the temperature coefficient. A well-trained upstream network $f(\cdot)$ can extract effective features and use them in the downstream network, which usually is a simple MLP network, mapping the latent vectors to different classes through supervised learning.

\section{Methodology}
\label{sec:methodology}
\subsection{Overview}
In this part, we illustrate the framework of our proposed data-free SCME in Fig. \ref{fig:framework}, which contains the following steps: 1) Synthesised Examples Generation and 2) Model Extraction. For step 1), we use a generator $\mathcal{G}(\cdot)$ to generate the fake data $\mathcal{X}$. In step 2), we input the $\mathcal{X}$ into both substitute model $\mathcal{F}_{sub}$ and target model $\mathcal{F}_{tgt}(\cdot)$ to minimize the difference of their outputs. Notably, $\mathcal{F}_{sub}(\cdot)$ in SCME consists of upstream encoder network $\mathcal{F}_{up}(\cdot)$, a feature extraction network, projector network ${F}_{down}(\cdot)$, and a classifier. Mathematically, the $\mathcal{F}_{sub}(\cdot)$ can be written as:
\begin{equation}
    \mathcal{F}_{sub}(x) = F_{down}(F_{up}(x)),
\end{equation}
where $x$ is an arbitrary input example.

Based on the two steps mentioned above, the ${F}_{sub}$ can imitate the $\mathcal{F}_{tgt}$ in a data-free manner. Finally, we can generate adversarial examples by attacking $\mathcal{F}_{sub}$, and further attack the target model $\mathcal{F}_{tgt}$ successfully.

\subsection{Intra- and Inter-class Diverse}
\label{subsec:Intra-class}
\begin{figure}[t!]
    \setlength{\abovecaptionskip}{-0.0cm}
    \setlength{\belowcaptionskip}{-0.5cm} 
    \includegraphics[width=\textwidth]{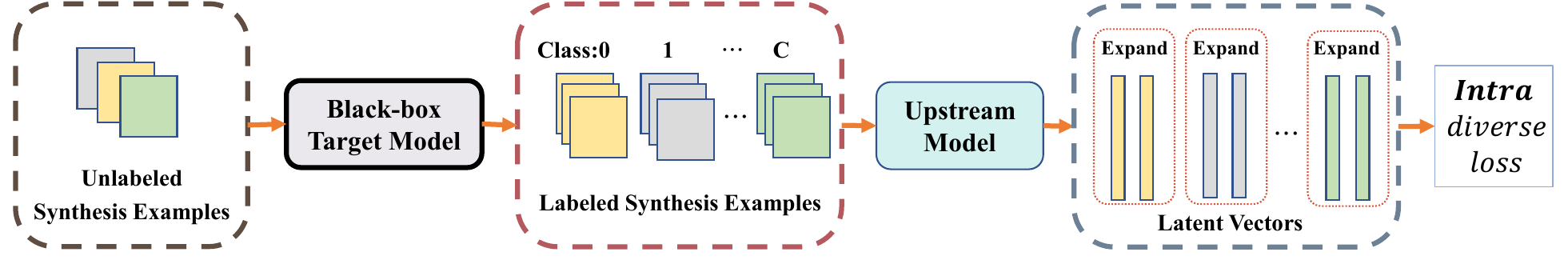}
    \caption{The calculation process of intra-class diverse loss. $C$ is the number of classes.}
    \label{fig:intra_diverse}
\end{figure}

As mentioned above, $\mathcal{X}$ should have both inter-class diversity and intra-class diversity to help the surrogate model training. Regarding this, as Fig. \ref{fig:intra_diverse} show, we propose a self-contrastive loss to guide $\mathcal{G}(\cdot)$ in the $\mathcal{X}$ generation. Inspired by the self-supervised loss in contrastive learning, we design a self-contrastive loss in SCME. Firstly, SCME uses the generator $\mathcal{G}(\cdot)$ to sample a batch of random noise ${N} = \{n_1, n_2, \cdot \cdot \cdot, n_B\}$ to generate corresponding synthesize examples $\mathcal{X} = \{x_1, x_2, \cdot \cdot \cdot, x_B\}$. SCME puts the $\mathcal{X}$ into the feature extraction network $\mathcal{F}_{up}(\cdot)$ and gets the latent vectors $z$. Then, SCME calculates the self-contrastive intra-class diverse loss $\mathcal{L}_{intra}$ by expanding the distance of each hidden vector $z_i$ in $z$. The self-contrastive loss can be formulated as:
\begin{equation}
    \mathcal{L}_{intra} = log\sum_{i}^{B}\sum_{j}^{B} \mathbbm{1}_{[i \neq j]} \cdot   exp(sim(z_i,z_j)),
    \label{eq:diverse}
\end{equation}
where $B$ is the batch size, $\mathbbm{1}$ is the indicator function and $sim(\cdot)$ is a similarity function. 

In the synthesised examples generation, the loss function $\mathcal{L}_{G}$ of generator $\mathcal{G}(\cdot)$ contains inter-class diversity loss $\mathcal{L}_{inter}$ and intra-class diversity loss $\mathcal{L}_{intra}$. To generate inter-class diversity examples, we use the inter-class information entropy to guide the generator $\mathcal{G}(\cdot)$. That is, SCME randomly sets a batch of target label $y_{tgt}$ and reduces the entropy between $y_{tgt}$ and the substitute model's output of the generated examples $\mathcal{X}$. Mathematically, the inter-class loss function is:
\begin{equation}
    \mathcal{L}_{inter} = \sum_{i=1}^{B}  \mathcal{F}_{sub}(\mathcal{X}_i)log[{F}_{sub}(\mathcal{X}_i)],
\end{equation}
where $B$ is the batch size.

\subsection{Model-independent Boundary Example}
Although the synthesis examples have been generated, however, they are still challenging to detect the target model's decision boundary adequately for substitute model training, resulting in a low attack performance. To solve this problem, we further modify the boundary examples by the Mixup augmentation to improve the substitute model's training efficiency. Specifically, SCME randomly selects two synthesized examples $\mathcal{X}_{i}$ and $\mathcal{X}_{j}$ first and then uses the Mixup to fuse them together to get the new boundary examples $\hat{\mathcal{X}}$, the process can be written as follows:
\begin{equation}
    \hat{\mathcal{X}} = \lambda \mathcal{X}_{i} + (1 - \lambda) \mathcal{X}_{j},
    \label{eq:mixup}
\end{equation}
where the $\lambda \in [0,1]$ is the mix weight and randomly sampled from $\beta$ distribution.

\subsection{Objective Function}
% \label{subsec:Inter-class}
By combining the above inter-class diversity loss $\mathcal{L}_{inter}$ and the intra-class diversity loss $\mathcal{L}_{intra}$, we obtain the generate loss $\mathcal{L}_G$ as the objective function for training the generator:
\begin{equation}
    \mathcal{L}_{G} = \mathcal{L}_{intra} + \alpha \mathcal{L}_{inter},
\end{equation}
where the $\alpha$ is the hyperparameter to adjust the weight of each loss.

Once the intra-class and inter-class diverse examples are generated, we input them into both substitute model $\mathcal{F}_{sub}(\cdot)$ and the target model $\mathcal{F}_{tgt}(\cdot)$ to minimize the distance between their outputs. To craft more suitable examples for training the substitute model and make its decision boundary close to the target model in the training process, we first craft the generated examples by Mixup operation to get the boundary examples $\hat{\mathcal{X}}$. Mathematically, the objective loss function $\mathcal{L}_{train}$ of training substitute model is:
\begin{equation}
    \mathcal{L}_{train} = \sum_{i=1}^{B}  d(\mathcal{F}_{sub}(\hat{\mathcal{X}}), \mathcal{F}_{tgt}(\hat{\mathcal{X}})),
\end{equation}
where the distance function $d(\cdot)$ is the Cross-Entropy loss in the hard label scenario and is the Mean Square Error loss in the soft label scenario.

Once the surrogate model is well-trained, we are able to generate adversarial examples on the substitute model and further attack the target black-box model.

\section{EXPERIMENTS}
\label{Sec:experiments}

\begin{table*}[]
    \setlength{\abovecaptionskip}{-0.3cm}
    \setlength{\belowcaptionskip}{-0.0cm} 
    \caption{Attack performance on MNIST and Fashion-MNIST Datasets.}
    \renewcommand{\arraystretch}{1.2}
    \small
    \centering
     \resizebox{\linewidth}{!}{
    
    \begin{tabular}{c|c|ccc|ccc|ccc|ccc} 
        \hline
        \multirow{2}{*}{Dataset}                                                 & \multirow{2}{*}{Methods} & \multicolumn{3}{c|}{Targeted, Hard Label}                            & \multicolumn{3}{c|}{Untargeted, Hard Label}      & \multicolumn{3}{c|}{Targeted, Soft Label}        & \multicolumn{3}{c}{Untargeted, Soft Label}        \\
        &                          & FGSM                     & BIM                      & PGD            & FGSM           & BIM            & PGD            & FGSM           & BIM            & PGD            & FGSM           & BIM            & PGD             \\ 
        \hline
        \multirow{6}{*}{\rotatebox{90}{MNIST}}                                                   & JPBA                     & 3.89                     & 6.89                     & 5.31           & 18.14          & 23.56          & 20.18          & 4.29           & 7.02           & 5.49           & 18.98          & 25.14          & 21.98           \\
        & Knockoff                 & 4.18                     & 6.03                     & 4.66           & 19.55          & 27.32          & 22.18          & 4.67           & 6.86           & 5.26           & 21.35          & 28.56          & 23.34           \\
        & DaST                     & 4.33                     & 6.49                     & 5.17           & 20.15          & 27.45          & 27.13          & 4.57           & 6.41           & 5.34           & 25.36          & 29.56          & 29.14           \\
        & Del                      & 6.45                     & 9.14                     & 6.13           & 22.13          & 25.69          & 23.18          & 6.97           & 9.67           & 6.24           & 24.56          & 25.35          & 25.28           \\
        & EBFA                     & \textbf{14.45}           & \textbf{28.71}           & 9.86           & 39.73          & 57.54          & 52.73          & \textbf{16.99} & \textbf{36.82} & \textbf{14.55} & 36.45          & 58.48          & 48.46           \\
        & \textbf{SCME}            & 9.98   & 9.96 & \textbf{10.04} & \textbf{63.45} & \textbf{74.51} & \textbf{78.47} & 10.05          & 10.00          & 10.04          & \textbf{72.46} & \textbf{78.54} & \textbf{82.54}  \\ 
        \hline
        \multirow{6}{*}{ \rotatebox{90}{Fashion-MNIST}} & JPBA                     & 6.45                     & 8.46                     & 7.57           & 24.22          & 30.56          & 30.11          & 6.89           & 8.56           & 7.56           & 26.23          & 31.35          & 31.11           \\
        & Knockoff                 & 6.34                     & 8.35                     & 7.32           & 28.19          & 36.88          & 35.92          & 6.65           & 8.98           & 8.23           & 30.21          & 36.94          & 36.22           \\
        & DaST                     & 5.38                     & 7.18                     & 6.53           & 30.45          & 36.17          & 34.23          & 5.33           & 7.46           & 7.84           & 32.14          & 37.34          & 34.91           \\
        & Del                      & 3.89                     & 8.19                     & 7.47           & 28.14          & 34.14          & 32.45          & 3.23           & 8.59           & 8.11           & 31.43          & 36.26          & 33.87           \\
        & EBFA                     & 30.08                    & \textbf{76.46}           & 32.42          & \textbf{84.85} & 80.93          & \textbf{89.30} & 29.11          & 66.02          & 43.56          & 75.19          & 79.94          & 79.30           \\
        & \textbf{SCME}            & \textbf{31.82}           & 70.79                    & \textbf{70.01} & 82.26          & \textbf{84.76} & 85.11          & \textbf{32.14} & \textbf{72.07} & \textbf{72.07} & \textbf{82.58} & \textbf{85.46} & \textbf{85.86}  \\
        \hline
    \end{tabular}
    }
    \label{table:mnist}
\end{table*}

\begin{table*}[ht]
    \setlength{\abovecaptionskip}{-0.4cm}
    \setlength{\belowcaptionskip}{-0.0cm} 
    \caption{Attack performance CIFAR-10 and CIFAR-100 Datasets.}
    \label{table:cifar}
    \renewcommand{\arraystretch}{1.2}
    \small
    \centering
     \resizebox{\linewidth}{!}{
    \begin{tabular}{c|c|ccc|ccc|ccc|ccc} 
        \hline
        \multirow{2}{*}{Dataset}  & \multirow{2}{*}{Methods} & \multicolumn{3}{c|}{Targeted, Hard Label}        & \multicolumn{3}{c|}{Untargeted, Hard Label}         & \multicolumn{3}{c|}{Targeted, Soft Label}        & \multicolumn{3}{c}{Untargeted, Soft Label}        \\
                                  &                          & FGSM           & BIM            & PGD            & FGSM            & BIM             & PGD             & FGSM           & BIM            & PGD            & FGSM           & BIM            & PGD             \\ 
        \hline
        \multirow{6}{*}{\rotatebox{90}{CIFAR-10}}  & JPBA                     & 6.32           & 7.70           & 7.92           & 27.82           & 33.23           & 31.70           & 7.28           & 8.56           & 7.64           & 28.77          & 33.38          & 31.96           \\
                                  & Knockoff                 & 6.26           & 7.02           & 7.04           & 29.61           & 31.86           & 30.68           & 6.46           & 8.27           & 7.35           & 30.02          & 31.98          & 30.35           \\
                                  & DaST                     & 6.54           & 7.81           & 7.41           & 27.61           & 34.43           & 26.99           & 8.15           & 8.40           & 8.26           & 27.58          & 34.75          & 27.47           \\
                                  & Del                      & 7.14           & 7.44           & 6.95           & 25.33           & 30.45           & 30.34           & 7.86           & 8.29           & 7.17           & 26.38          & 31.53          & 31.47           \\
                                  & EBFA                     & 14.57          & \textbf{16.95} & 12.27          & 86.13           & 87.02           & 84.32           & \textbf{31.54} & 13.93          & \textbf{69.14} & 83.89          & 87.68          & 85.11           \\
                                  & \textbf{SCME}            & \textbf{16.53} & 14.76          & \textbf{14.22} & \textbf{91.01 } & \textbf{91.56 } & \textbf{91.33 } & 16.22          & \textbf{15.14} & 15.31          & \textbf{91.23} & \textbf{91.62} & \textbf{91.66}  \\ 
        \hline
        \multirow{6}{*}{\rotatebox{90}{CIFAR-100}} & JPBA                     & 4.35           & 6.20           & 6.17           & 33.58           & 38.54           & 37.08           & 5.73           & 7.50           & 6.41           & 34.21          & 39.12          & 37.31           \\
                                  & Knockoff                 & 4.40           & 5.86           & 5.25           & 34.84           & 36.92           & 36.34           & 4.88           & 7.05           & 6.18           & 36.01          & 37.61          & 35.47           \\
                                  & DaST                     & 4.97           & 6.19           & 5.92           & 33.57           & 39.86           & 32.71           & 6.38           & 7.04           & 7.01           & 32.80          & 40.34          & 32.78           \\
                                  & Del                      & 5.38           & 5.72           & 5.69           & 30.80           & 35.63           & 36.15           & 6.30           & 6.53           & 5.23           & 31.64          & 36.63          & 37.44           \\
                                  & EBFA                     & 16.64          & \textbf{16.88} & 12.77          & 78.61           & 91.31           & 91.21           & 7.91           & \textbf{16.15} & 12.54          & 83.69          & 94.53          & 94.14           \\
                                  & \textbf{SCME}            & \textbf{18.46} & 14.23          & \textbf{13.13} & \textbf{94.81}  & \textbf{95.40}  & \textbf{95.32}  & \textbf{10.50}  & 16.02          & \textbf{15.49} & \textbf{94.72} & \textbf{95.09} & \textbf{94.95}  \\
        \hline
    \end{tabular}
    }
    \vspace{-20pt}
\end{table*}

\subsection{Setup}
\label{subsec:setup}
\textbf{Datasets}: We consider five benchmark datasets, namely MNIST \cite{mnist}, Fashion-MNIST \cite{Fashion}, CIFAR-10 \cite{cifar10}, CIFAR-100 \cite{cifar10}, Tiny-ImageNet \cite{Tiny} for comprehensive experiments. 

\textbf{Models}: For MNIST and Fashion-MNIST datasets, we use a simple network as the target model, which has four convolution layers and pooling layers and two fully-connected layers. For CIFAR-10 and CIFAR-100, we use the ResNet-18 \cite{Resnet} as the target model. For Tiny-ImageNet, we use the ResNet-50 \cite{Resnet} as the target model. The substitute model for all the datasets is the VGG-16 \cite{VGGnet}.

\textbf{Baselines}: To evaluate the performance of SCME, we compare it with the data-dependent method, JPBA \cite{DBLP:conf/ccs/PapernotMGJCS17}, Knockoff \cite{Knockoff}, and data-free methods, DAST \cite{DAST}, Del \cite{Del}, EBFA \cite{EBFA}.

\textbf{Training details}: SCME and the baseline methods are trained with Adam optimizer with batch size 256. For the generator in SCME, we use an initial learning rate of 0.001 and a momentum of 0.9, and for the substitute model, we set the initial learning rate as 0.01 and momentum as 0.9. Furthermore, we set the maximal query times as 20K, 100K and 250K for the MNIST dataset, CIFAR dataset and the Tiny-ImageNet dataset, respectively.

\begin{table}[ht]
    \setlength{\abovecaptionskip}{-0.5cm}
    \setlength{\belowcaptionskip}{-0.0cm} 
    \caption{Attack performance on Tiny-ImageNet Dataset.}
    \renewcommand{\arraystretch}{1.2}
    \centering
    \begin{tabular}{c|ccc|ccc}
        \hline
        \multicolumn{1}{l|}{} & \multicolumn{3}{c|}{Hard Label} & \multicolumn{3}{c}{Soft Label} \\ \hline
        \multicolumn{1}{l|}{Methods} & FGSM & BIM & PGD & FGSM & BIM & PGD \\ \hline
        JPBA & 15.37 & 25.16 & 14.23 & 26.54 & 28.91 & 26.83 \\
        Knockoff & 22.33 & 21.39 & 11.26 & 29.99 & 27.64 & 26.17 \\
        DaST & 16.23 & 18.26 & 15.86 & 28.81 & 29.37 & 26.51 \\
        Del & 28.31 & 32.54 & 29.73 & 34.28 & 38.49 & 36.72 \\
        EBFA & 78.29 & 81.12 & 78.23 & 80.26 & 85.32 & 78.29 \\
        SCME & \textbf{90.16} & \textbf{90.25} & \textbf{89.72} & \textbf{96.44} & \textbf{96.29} & \textbf{96.32} \\ \hline
    \end{tabular}
    \label{tiny}
    % \vspace{-15pt}
\end{table}

\begin{figure}[htbp]
    \setlength{\abovecaptionskip}{-0.1cm}
    \setlength{\belowcaptionskip}{-0.0cm} 
    \begin{minipage}[t]{0.5\linewidth}
        \centering
        \includegraphics[width=0.95\textwidth]{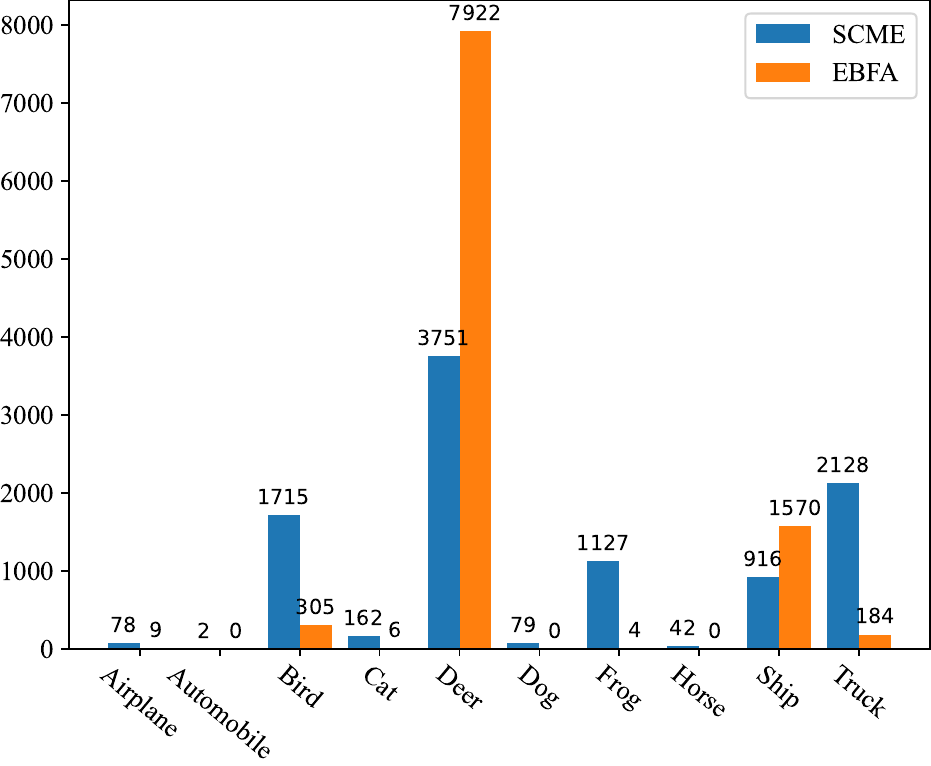}        
    \end{minipage}%
    \begin{minipage}[t]{0.5\linewidth}
        \centering
        \includegraphics[width=0.95\textwidth]{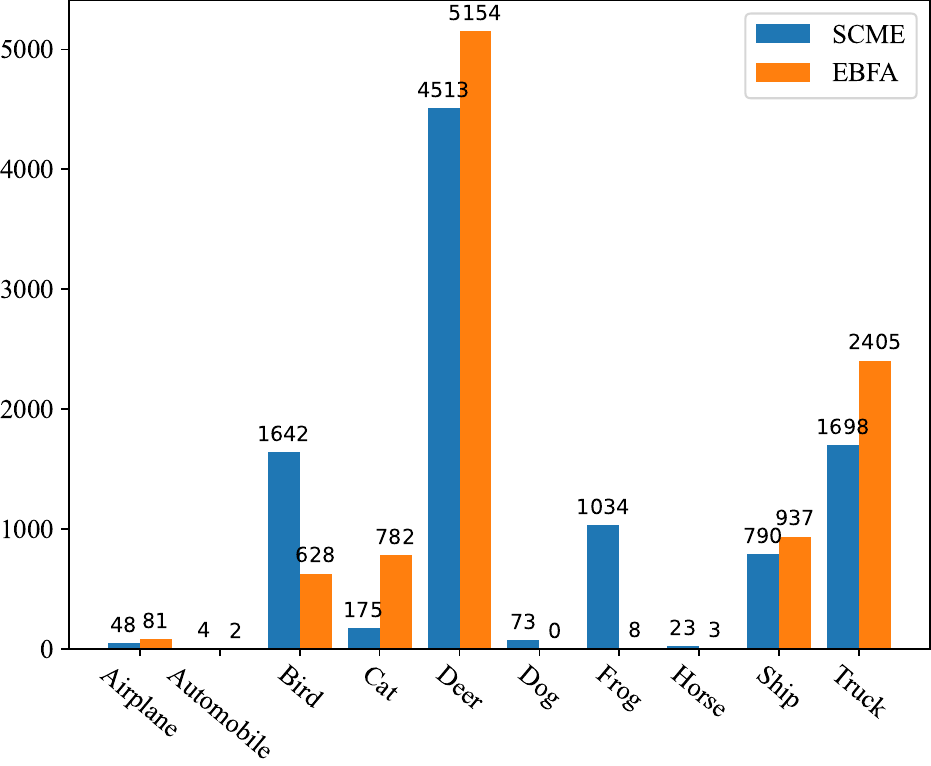}   
    \end{minipage}
    \caption{Synthesised examples without (left) and with (right) data augmentation.}
    \label{fig:data-aug}
    \vspace{-15pt}
\end{figure}

\textbf{Metrics}: We utilize three classical attack methods, which include FGSM, BIM and PGD, to generate adversarial examples for the surrogate model. For MNIST and Fashion-MNIST, we set perturbation budget $\epsilon$ = 32/255. And for CIFAR-10, CIFAR-100 and Tiny-ImageNet, we set $\epsilon$ = 8/255. In the untargeted attack scenario, we only generate adversarial examples for the images which can be classified correctly by the victim model, while in targeted attacks, we only generate adversarial examples for the images which are not classified to the specific wrong labels. The \textbf{attack success rate} (ASR) is calculated by:
\begin{equation}
    ASR = \begin{cases}
    \frac{1}{N} \sum_{i=1}^{N}[f(x_{i}^{adv})\neq y_{i}], \ for \ untargeted \\
    \frac{1}{N} \sum_{i=1}^{N}[f(x_{i}^{adv}) = y_{t}], \ for \ targeted
    \end{cases} 
\end{equation}
where $N$ is the total number of generated adversarial examples. 

Besides, for given a batch of query examples $\mathcal{X}$, we input them to the target model $\mathcal{F}_{tgt}$ to get the output of each example and calculate its \textbf{Boundary Values} (BV) to verify whether the query samples are close to the decision boundary of the target model or not. The proposed BV can be calculated as follows:
\begin{equation}
    BV = \sum_{i=1}^{B}(p (\mathcal{F}_{tgt}(\mathcal{X}_i))_{top_1}-p(\mathcal{F}_{tgt}(\mathcal{X}_i))_{top_2}),
\end{equation}
where $p$ is the Soft-max function, the $top_1$ and $top_2$ are the maximum value and sub-maximal value in the output probability vector, and the $B$ is the total example counts.

\subsection{Attack Performance}
\textbf{Experiments on MNIST and Fashion-MNIST:} We report the ASR under targeted and untargeted attacks for both label-only and probability-only scenarios. As shown in Table~\ref{table:mnist}, the ASR of SCME is much higher than the SOTA baselines on MNIST and Fashion-MNIST datasets. Obviously, our method can obtain higher ASR than other baselines in most cases with a small number of queries (here is 20K). This phenomenon shows that the proposed method is more applicable to the real world than the baselines.

\begin{figure}[ht]
\vspace{-15pt}
    \setlength{\abovecaptionskip}{-5pt}
    \setlength{\belowcaptionskip}{-15pt} 
    \centerline{ 
        \includegraphics[width=0.3\textwidth]{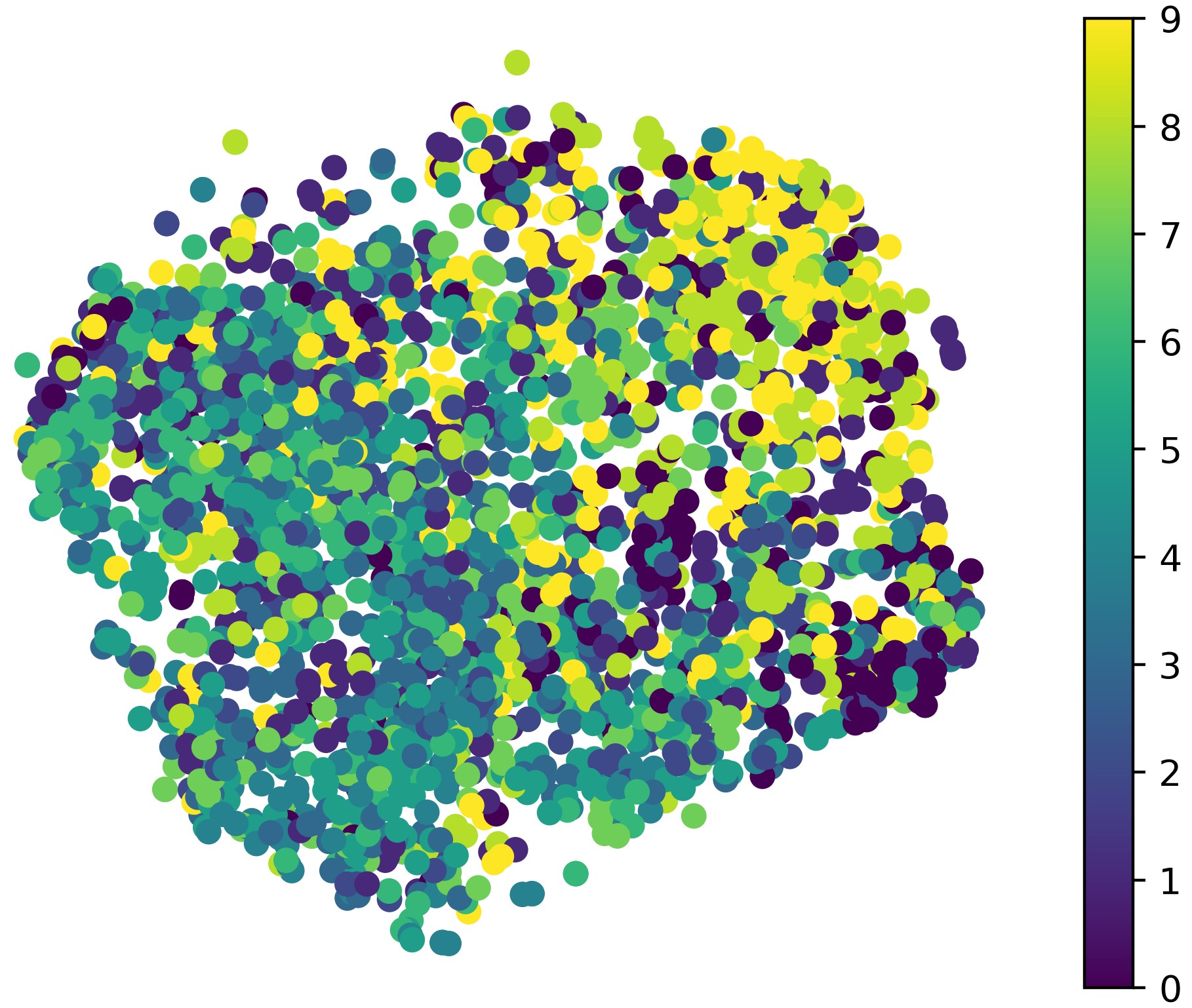}
         \includegraphics[width=0.3\textwidth]{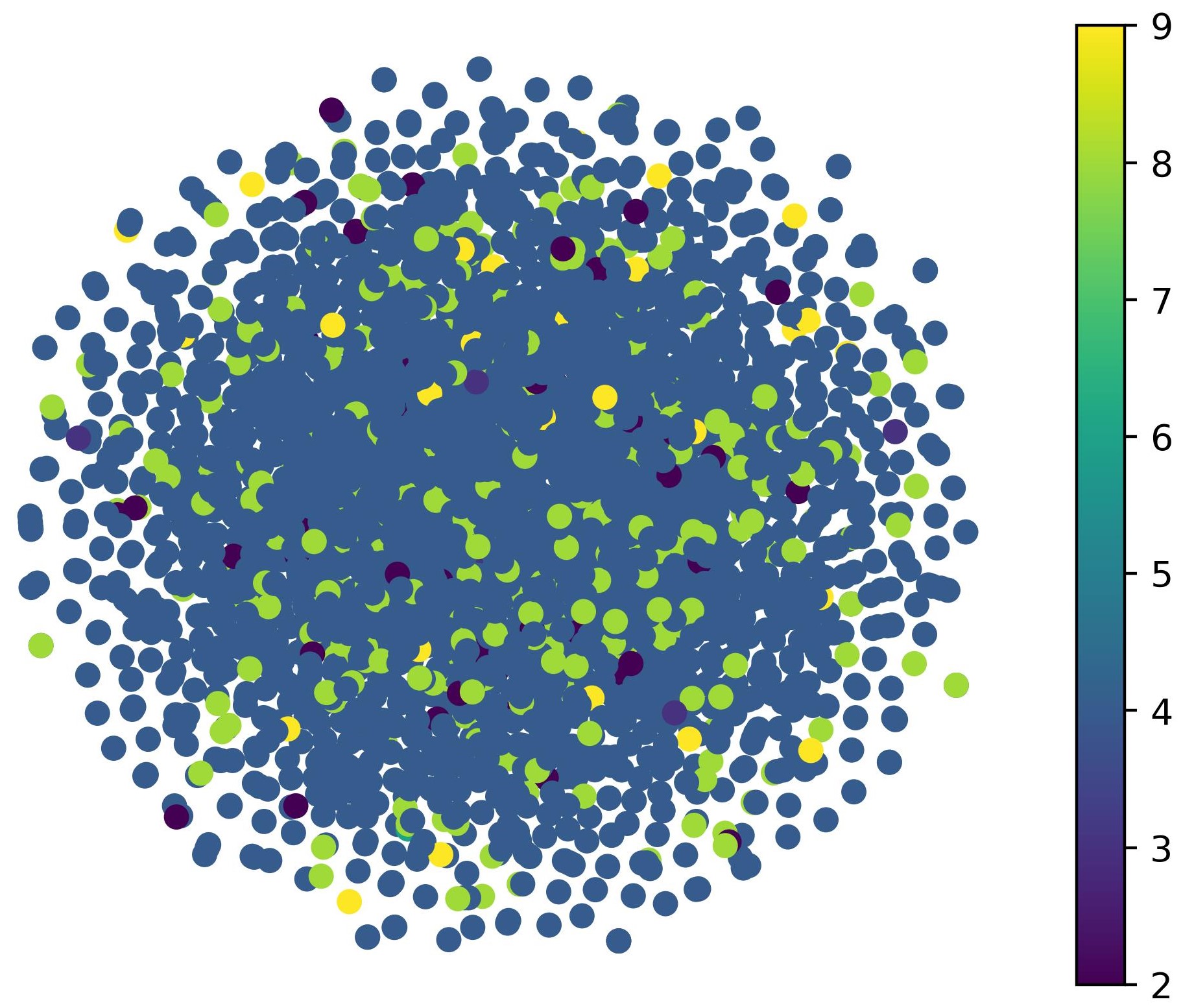}
         \includegraphics[width=0.3\textwidth]{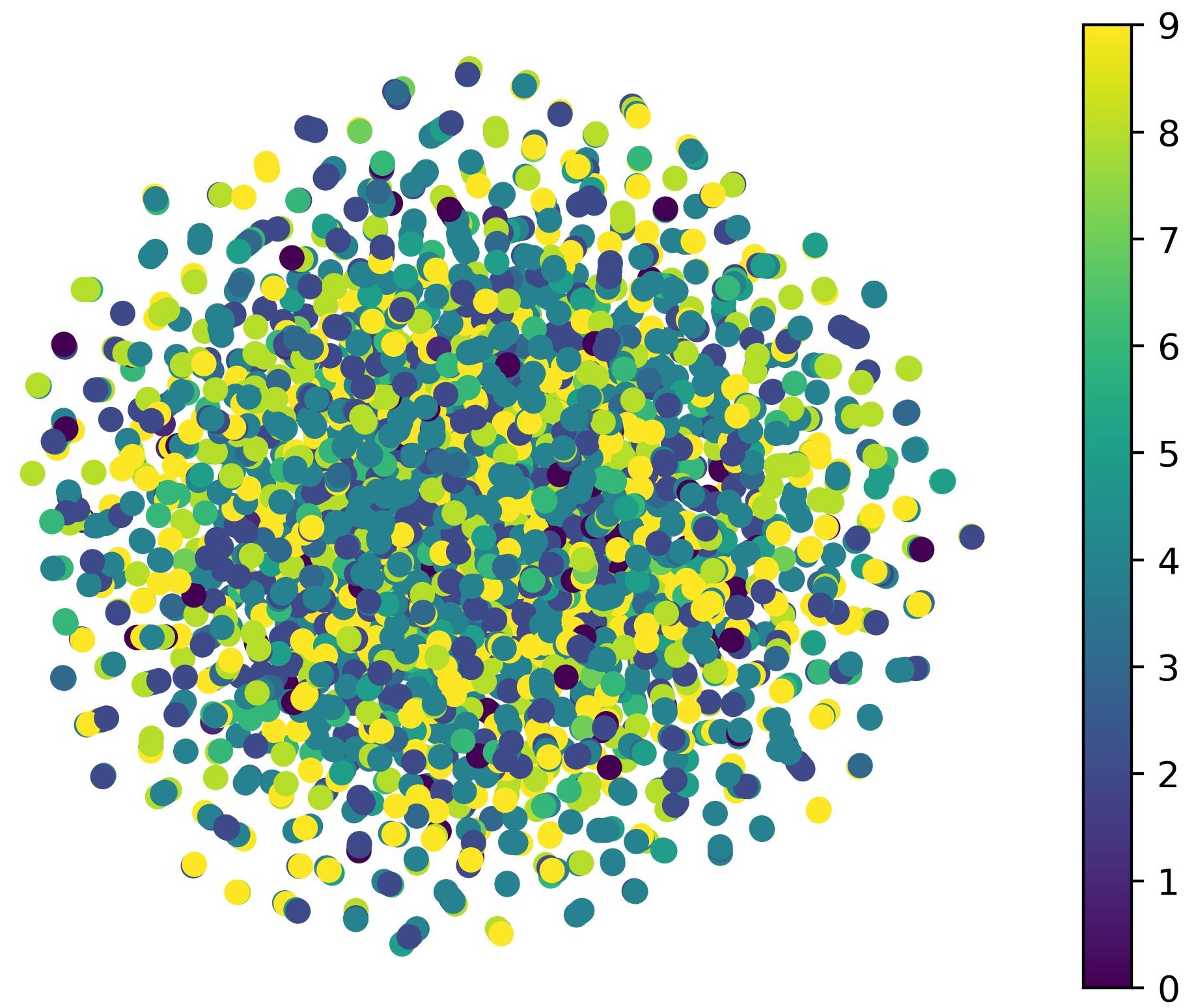}
     }
    \caption{The T-SNE of original CFIAR-10 data (left), synthetic data by EBFA (middle) and synthetic data by SCME (right).}
    \label{fig:tsne}
\end{figure}

\textbf{Experiments on CIFAR-10, CIFAR-100 and Tiny-ImageNet:} We further investigate the performance of our method on complex datasets. From the results shown in Tables~\ref{table:cifar} and~\ref{tiny}, our method achieves the best attack performance over probability-only and label-only scenarios under all datasets. In addition, compared to the strong baselines EBFA, our method still outperforms it significantly. Although the number of categories directly affects the training of the substitute model, our method still achieves a very high ASR on the CIFAR-100 and Tiny-ImageNet datasets, which have 100 categories and 200 categories, respectively. On the Tiny-ImageNet dataset, our method even achieves the highest ASR of 96.44\% in the soft label setting. These improvements effectively demonstrate the superiority of the proposed SCME.

\begin{table}[ht]
\vspace{-10pt}
    \setlength{\abovecaptionskip}{-0.0cm}
    \setlength{\belowcaptionskip}{-0.0cm} 
    \caption{Boundary value of EBFA and SCME.}
    \centering
    \renewcommand\arraystretch{1.2}
    \begin{tabular}{c|cc|cc} 
    \hline
    \multirow{2}{*}{Methods} & \multicolumn{2}{c|}{EBFA}           & \multicolumn{2}{c}{SCME}               \\
                             & w.o. aug. & w. aug. & w.o. aug. & w. aug.     \\ 
    \hline
    Boundary Values          & \textbf{9150.8699}       & 9010.6072         & 9469.6909       & \textbf{8717.2107}  \\
    \hline
    \end{tabular}
    \label{table:boundaryvalue}
    \vspace{-25pt}
\end{table}

\begin{figure}[!ht]
    \setlength{\abovecaptionskip}{-0.0cm}
    \setlength{\belowcaptionskip}{-0.0cm} 
    \centering
    \includegraphics[width=0.47\textwidth]{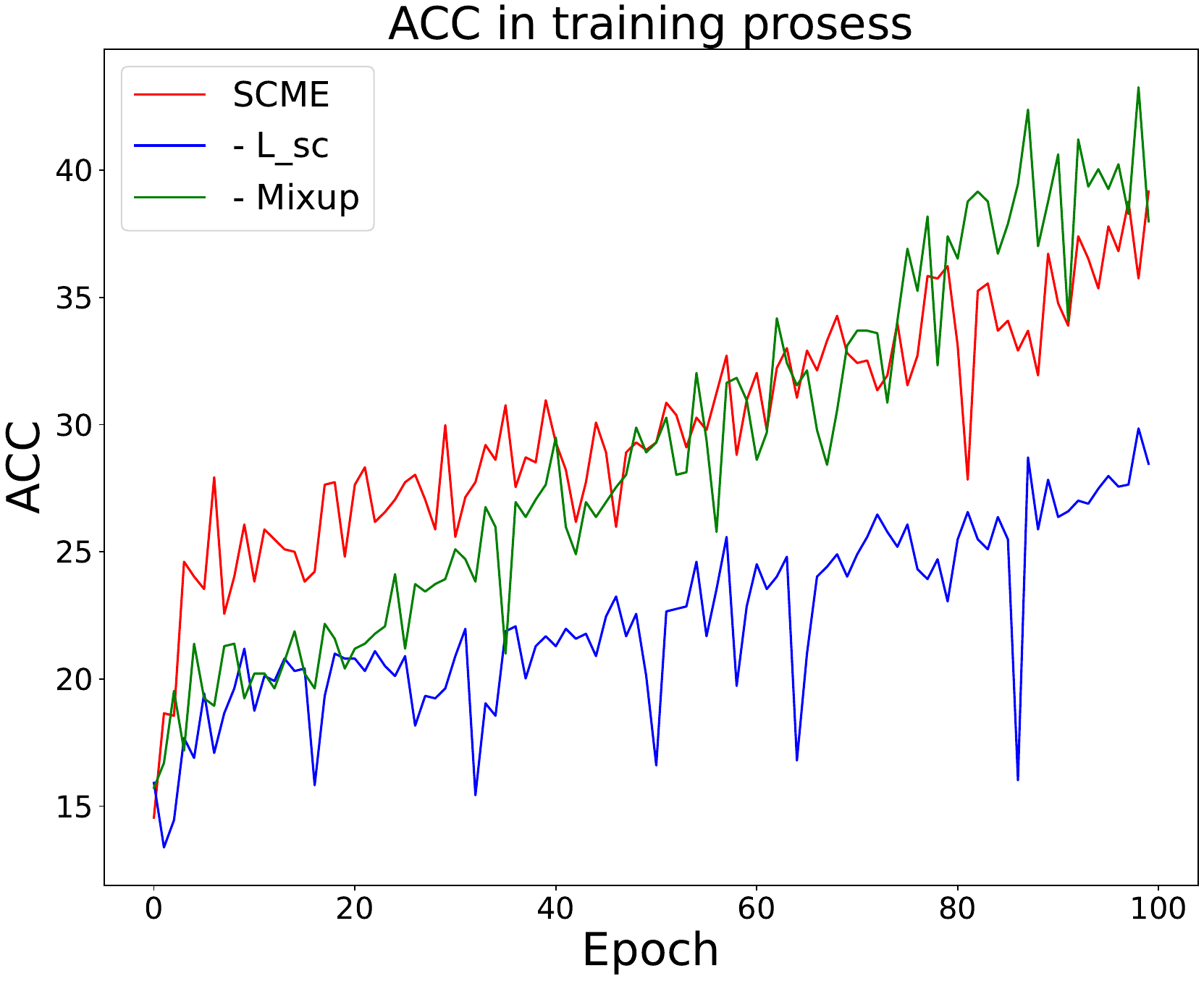}
    \hspace{0.1in}
    \includegraphics[width=0.47\textwidth]{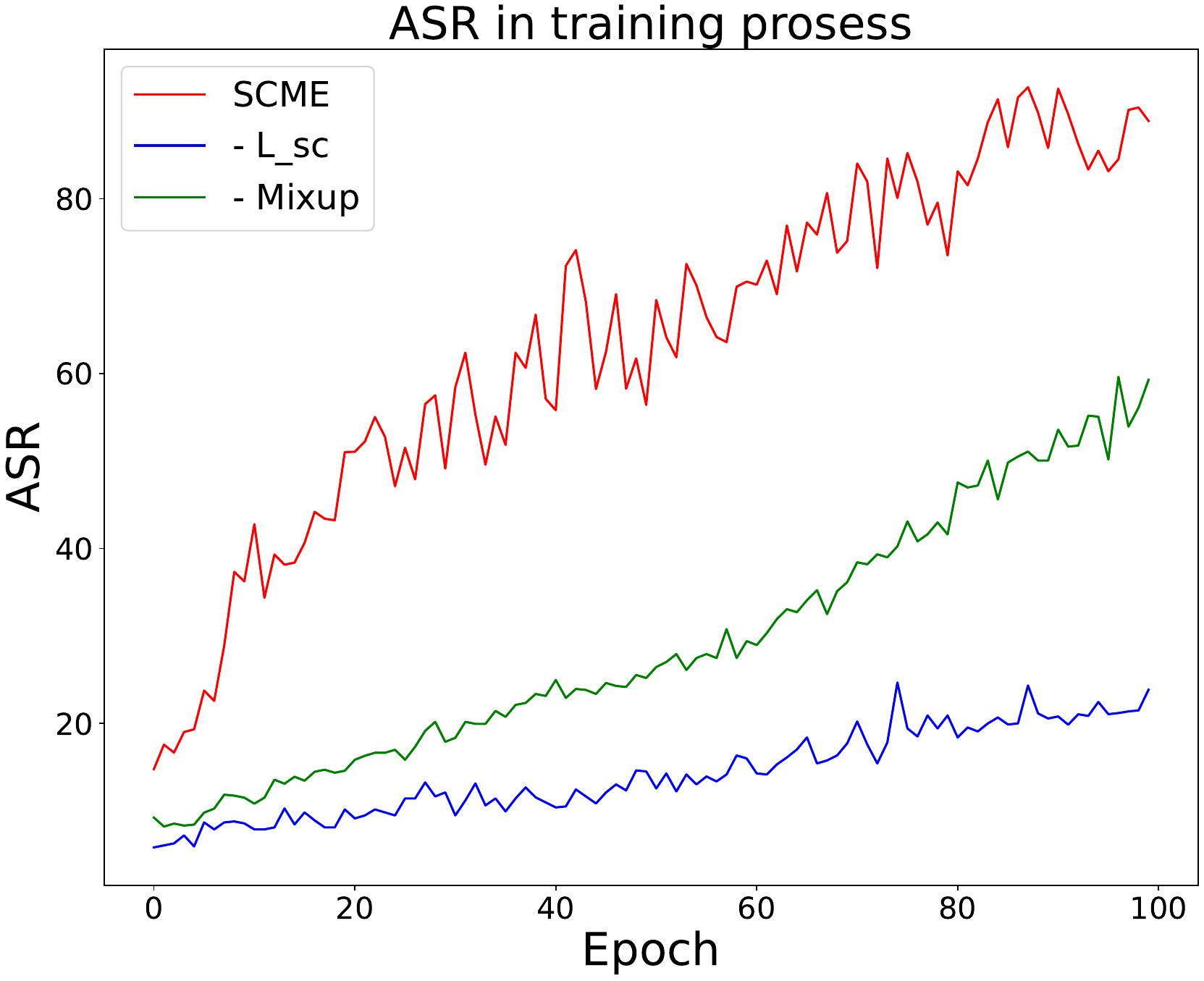}
    \caption{The ablation results of model accuracy (left) and model ASR (right), where "- L\_sc" means without self-contrastive loss, and "- Mixup" means without Mixup operation.}
    \label{fig:ablation}
    \vspace{-10pt}
\end{figure}

\subsection{Evaluation on Data Diversity}
To evaluate the generated data's diversity of the strong baseline EBFA and the proposed SCME, we generated 10,000 examples and fed them into the same model trained on the CIFAR-10 dataset to get the predicted labels. The results in Fig. \ref{fig:data-aug} show the data with data augmentation or not. The results show most of the examples generated by EBFA were classified as "deer", while synthetic examples by SCME have preferable inter-class diversity. Further, we plot the T-SNE for real data and synthetic examples generated by EBFA and SCME, respectively, in Fig. \ref{fig:tsne}. The results illustrated that our method generates examples similar to the real data, i.e., with more intra-class diversity. These phenomena strongly support that our method can generate data with high inter- and intra- diversity. 

\subsection{Evaluation on Boundary Value}
To verify whether the query examples are closer to the decision boundary of the target model, we compared the BV of 10,000 examples generated by EBFA and SCME. The results in Table~\ref{table:boundaryvalue} show although EBFA achieves smaller BV without data augmentation, SCME can achieve substantially lower BV with data augmentation. This further demonstrates the effectiveness of the Mixup operation in SCME for generating query examples close to the decision boundary.

\subsection{Ablation Study}
To investigate the contribution of Self-Contrastive loss $\mathcal{L}_{G}$ (described in Sec. \ref{subsec:Intra-class} and Mixup operation, we plot the model classification accuracy (ACC) and the model ASR in the model training process. The results in Fig. \ref{fig:ablation} shows that using both $\mathcal{L}_{G}$ and Mixup augmentation performs best on both ACC and ASR, besides the model training convergence faster. For instance, the standard SCME is close to convergence with 6K queries, and the ASR is also beyond 80\%.

\section{Conclusion}
\label{Sec:conclusion}
In this paper, we proposed a novel data-free model extraction attack, namely SCME, to boost the attack performance under query-limited settings. Specifically, we first design a self-contrastive loss to guide the generator to synthesize the query data with high inter- and intra-class diversity. Besides, we introduce the Mixup augmentation to combine two generated query samples as the final query input to obtain effective decision boundaries and further help the simulation process of the substitute model. Extensive empirical results show that the proposed SCME framework can achieve SOTA attack performance.

\subsubsection{Acknowledgements} 
This work is supported in part by Yunnan Province Education Department Foundation under Grant No.2022j0008, in part by the National Natural Science Foundation of China under Grant 62162067 and 62101480, Research and Application of Object Detection based on Artificial Intelligence, in part by the Yunnan Province expert workstations under Grant 202205AF150145.

\bibliographystyle{splncs04}
\bibliography{ref}

\end{document}